%% file: root.tex
\documentclass[letterpaper, 10 pt, conference]{ieeeconf}  

\IEEEoverridecommandlockouts                              
\usepackage{cite}
\usepackage{amsmath}
\usepackage{amssymb}
\usepackage{amsfonts}
\usepackage{algorithmic}
\usepackage[ruled,linesnumbered,vlined]{algorithm2e}
\usepackage[final]{graphicx}
\usepackage{textcomp}
\usepackage{xcolor}
\usepackage{etoolbox}
\usepackage{comment}
\usepackage{cleveref}

\usepackage{mathtools}
\title{\LARGE \bf
Modelling Human Kinetics and Kinematics during Walking using Reinforcement Learning
}

\author{Visak Kumar
\thanks{}
\thanks{}%
}

\begin{document}

\maketitle
\thispagestyle{empty}
\pagestyle{empty}

\begin{abstract}
In this work, we develop an automated method to generate 3D human walking motion in simulation which is comparable to real-world human motion. At the core, our work leverages the ability of deep reinforcement learning methods to learn high-dimensional motor skills while being robust to variations in the environment dynamics. Our approach iterates between policy learning and parameter identification to match the real-world bio-mechanical human data. We present a thorough evaluation of the kinematics, kinetics and ground reaction forces generated by our learned virtual human agent. We also show that the method generalizes well across human-subjects with different kinematic structure and gait-characteristics.

\end{abstract}

\input{defs}
\section{INTRODUCTION}

Assistive devices such as exoskeletons have garnered a lot of research interest because they can bring numerous benefits to injury/disability rehabilitation, elderly care, human-augmentation for people doing heavy lifting at work, exercise and military applications. However, there a few challenges we need to overcome before assistive devices can become more prevalent in our society. Exoskeletons work by directly applying forces on the joints, hence safety during operation becomes paramount when designing control policies for such devices. Often, testing directly on a human-subject is not recommended.

Simulation has played a key role in offloading the burden of testing control policies in the real-world with human subjects. However, simulating the real-world accurately has been a long-standing challenge in robotics.  Modelling human motion is especially difficult because of the high-dimensional nature of the state space and the natural variability that exists between different human-subjects.

Our focus is on modelling human motion during locomotion tasks such as walking and running. Humanoid locomotion is governed by highly non-linear and discontinuous dynamics and large number of unknown quantities such as joint damping, segment Center of Mass (COM), segment inertia, friction coefficient, muscle activation and muscle-tendon interaction. These are extremely challenging to measure. Traditional methods to identify simulation models, commonly known as system identification, are often insufficient when the system in question is high-dimensional. System identification works by collecting data on the real-system and then inferring model parameters parameters based on the data collected. For a high-dimensional system, collecting data in the task-relevant state space can be quite challenging.  

The difficulty in modelling human locomotion has led some researchers to focus on reduced-order models, such as Linear-inverted pendulum models (LIPM) (or even 3D LIPM) \cite{Joshi2019}, to explain human locomotion. Although computational more efficient, these models come at the cost of accuracy and hence are limited in their application. 

Reinforcement learning (RL) and imitation learning have taken center stage in recent years in developing control algorithms to imitate human motion. In particular, Peng et al \cite{peng2018deepmimic} and  Yu et al \cite{Yu_2018} learn complex motor skills like walking, running, jumping and even performing a backflip. In addition to learning complex motion, policies trained using RL are also robust to variations in the underlying dynamics : variations caused by different dynamical parameters \cite{Yusimtoreal} as well as shape and size of the human-subject \cite{Won:2019}. 

In this work, we leverage the following two advantages of RL, \textbf{(1)} Ability to learn complex motions in high-dimensional space and \textbf{(2)} robustness to variability in underlying dynamics, to learn control policies for human walking that are bio-mechanically accurate. We validate our approach by comparing the joint angles, joint moments and ground reaction forces generated by the policy to real-world data collected in human-subject experiments.

\section{Related Work }
Since this work aims to leverage tools and data which are outcomes of research in two different fields - Deep Reinforcement Learning (DRL) and Biomechanics. The literature review is organized into two sections, each describes the current state of the art in each field and what aspects needs improvement. 

\subsection{Deep Reinforcement Learning}
DRL has seen remarkable success in recent years for learning complex tasks ranging from video games to controlling robots \cite{schulman2015trust,schulman2017proximal,DBLP:journals/corr/HasseltGS15}. In particular, DRL has been successful in the field of developing locomotion controllers \cite{peng2017deeploco,YuSIGGRAPH2018,Yusimtoreal,peng2018deepmimic}. However, these methods are seldom validated by comparing it to real human movement
data generated by experiments. In Peng et al \cite{peng2018deepmimic}, a policy optimized using DRL algorithm was able to learn exceptional skills like walking, running , jumping and even doing a backflip with the help of motion capture data. But, the resulting policy, while visually pleasing, is not validated with real-world human data. Similarly, in Yu et al \cite{YuSIGGRAPH2018}, several locomotion skills like walking and running at different speeds was learned from scratch using a curriculum learning approach. However, as was the issue with \cite{peng2018deepmimic}, this approach is not validated with real data. Additionally, the above mentioned approach only adopts symmetry and low energy as metrics to enforce in the learned walking strategy, but some research in biomechanics points towards an existence of more complex relationships that gives rise to walking. For example, Wang et al \cite{Wang2014} identified that the Center-of-mass velocity has a direct linear relationship to step-length. In Kumar et al \cite{kumar2019learning}, preliminary comparison of the walking motion learned using policy optimization and real-world human subject experiments were made. However, only joint kinematics and foot-step lengths were used as a metric for validation. We need a more thorough evaluation using joint kinematics, kinetics, ground reaction forces and foot step lengths across different individuals to be more confident about the simulation model.

\subsection{Biomechanics} 

To study human motion during walking and recovery, biomechanics researchers often adopt an experimental approach. First, data is collected in the real-world, then control policies are synthesized using the real-world data. Winters et al \cite{Winter} was among the first in the field to study human gait, and the data published in this work remains relevant to this day. Wang et al \cite{Wang2014} and Hof et al \cite{hof2010balance} performed perturbation experiments and identified important relationships between COM velocity, step-lengths, center of pressure, stepping vs ankle strategy, etc.. We aim to leverage the finding of this research to validate some of the results. In Joshi et al \cite{Joshi2019}, a balance recovery controller was derived using the results reported in \cite{Wang2014}, however, they use a 3D Linear Inverted pendulum model to approximate the human dynamics. A 3D LIPD does not capture the dynamics fully, for example, angular momentum about the center of mass. Most relevant to our work, Antoine et al \cite{antoine2019}, used a direct-collocation trajectory optimization to synthesize a walking controller for a 3D musculo-skeletal model in OpenSim (OpenSim gait2392 model). The gait generated by the controller closely matched experimental data. The proposed method, relies on understanding the basic principles that lead to walking, such as minimizing metabolic cost, muscle activations,etc. However, the proposed solution enforces left-right symmetry, which works for walking, but is not ideal for disturbance recovery. Hence its unclear how well this approach will perform when there is an external disturbance to the human.

\section{Method}
\label{sec:method}
We propose a framework to automate the process of developing bio-mechanically accurate 3D Human walking policies in simulation. The difference between the dynamics of a simulated human agent and a real human subject is caused by multiple factors. Some quantities such as mass, height, etc.. can be easily measured. However other quantities such as dynamical parameters: joint damping, ground friction, lower-limb joint axis location and the accurate segment lengths can be challenging to estimate. Note that while the segment lengths can be easily measured on a real human, the bio mechanical data sets usually do not provide this information.

We present an iterative approach to develop accurate walking model in which, first, we use Deep Reinforcement Learning (DRL) to learn a walking policy using a nominal dynamical model. During training, domain randomization is used to ensure the policy is robust to dynamical and kinematic parameters. Second, once we have a policy to generate walking motion, we perform an optimization step to identify the optimal parameters that explain real-world walking motion, these two steps are repeated until convergence.
To validate our simulated models, we compare the gait characteristics such as joint kinematics, kinetics and ground reaction forces generated by our policy to real-world data collected with 5 human participants.

We formulate this problem of learning human walking as a Markov Decision Processes (MDPs), $(\mathcal{S}, \mathcal{A}, \mathcal{T}, r, p_0, \gamma)$, where $\mathcal{S}$ is the state space, $\mathcal{A}$ is the action space, $\mathcal{T}$ is the transition function, $r$ is the reward function, $p_0$ is the initial state distribution and $\gamma$ is a discount factor. We take the approach of model-free reinforcement learning to find a policy $\pi$, such that it maximizes the accumulated reward:
\begin{equation}
    J(\pi) = \mathbb{E}_{\mathbf{s}_0, \mathbf{a}_0, \dots, \mathbf{s}_T} \sum_{t=0}^{T} \gamma^t r(\mathbf{s}_t, \mathbf{a}_t),\nonumber
\end{equation}
 where $\mathbf{s}_0 \sim p_0$, $\mathbf{a}_t \sim \pi(\mathbf{s}_t)$ and $\mathbf{s}_{t+1}=\mathcal{T}(\mathbf{s}_t, \mathbf{a}_t)$.

We denote the human walking policy as $\pi_{h}(\vc{a}_{h}|\vc{s}_{h})$ where $\vc{s}_h$, $\vc{a}_h$  represent the states and actions, respectively.

\section{Data and Simulation Environment}

Fukuchi et al \cite{Fukuchi2018} published open-source biomechanics data of 38 human subjects walking overground and on treadmill. We use data of 10 subjects (for proof-of-concept) with different physical characteristics such as mass, height, leg-length, gender and speed of walking. For each subject, this dataset provides motion capture marker data, and individual gait characteristics like joint angles, joint moments and ground reaction forces. 

We use DART physics \cite{pydart} engine as our simulation environment in which the virtual human agent learns to walk. We first create 10 virtual agents with the same physical characteristics as the human subjects shown in table \ref{tab:data}. We choose hunt-crossley contact model to generate contact forces between two rigid bodies such as feet and ground. Hunt-crossley model is widely used in biomechanics research and is also one of the contact models used in OpenSim simulator \cite{OpenSim}.

\begin{table}[!htb]
\centering

\begin{tabular}{|l|l|l|l|l|}
\hline
 Subject&Mass (kg) &Height (cm)  & Age  & Speed (m/s)\\ \hline
 1 (M)&74  &172.40  &25  &1.25  \\ \hline
 2 (M) &52.9  & 166.80  &22  & 1.40  \\ \hline
 3 (F)& 48.8  &158  &24  &1.15  \\ \hline
 4 (M)& 61.5  & 180.70  & 22  & 1.28  \\ \hline
 5 (F)&153  &64  &31  &1.09  \\ \hline
 6 (M)& 69.85  &155  &38  &1.3  \\ \hline
 7 (F)&64.6  &151.50  &57  &0.91  \\ \hline
 8 (M)&63.3  &175  &71  &0.57  \\ \hline
 9 (F)&46.05  &149.20  &63  &0.80  \\ \hline
 10 (M)&66.35  &155.50  & 84 &0.60  \\ \hline
\end{tabular}
\caption{Human subject physical characteristics.}
\label{tab:data}
\end{table}

\begin{figure}[!htb]
\centering
\setlength{\tabcolsep}{1pt}
\renewcommand{\arraystretch}{0.7}
 
  \includegraphics[width=0.19\textwidth,height=3cm]{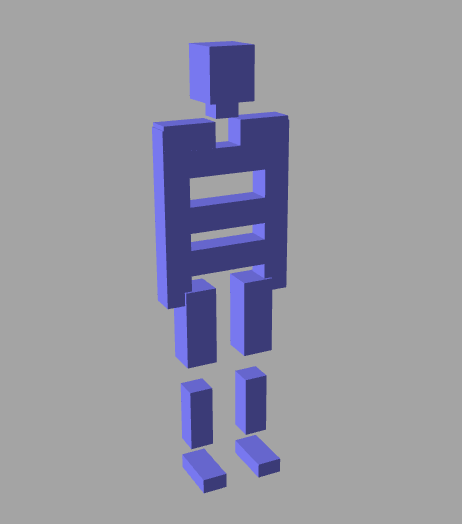}

\caption{ We model a 31-Degree of Freedom(DoF) humanoid  in PyDart. }
\label{fig:Dart}
\end{figure}






\subsection{Human Walking Policy}
\label{sec:human_policy}
We take a model-free reinforcement learning approach to develop a human locomotion policy $\pi_{h}(\vc{a}_{h}|\vc{s}_{h})$. To achieve natural walking behaviors, we train a policy that imitates the human walking reference motion (motion capture data from the open-source data base) similar to Peng \etal \cite{peng2018deepmimic}. The human 3D model (agent) consists of $25$ actuated joints with a floating base as shown in \figref{Dart}. This gives rise to a $71$ dimensional state space $\vc{s}_{h} = [\vc{q},\vc{\dot{q}},\vc{v}_{com}, \boldsymbol{\omega}_{com},\phi,\mu]$, including joint positions, joint velocities, linear and angular velocities of the center of mass (COM), and a phase variable($\phi$) that indicates the target frame in the motion clip. Vector $\mu$ ($R^14$) described as $$\mu\ =\ [\beta,\sigma,f_l,s_l,t_l,f_j,h_j,s_j]$$ 
where $\beta$ is joint damping, $\sigma$ is friction coefficient between the foot and the ground. $f_l,s_l,t_l$ and foot, shin and thigh segment lengths respectively and $f_j,h_j,s_j$ are their corresponding joint axis location expressed in the robot root coordinate frame. These quantities are randomized during training so that the learned policy can be robust to variations in these quantities. 

The action determines the target joint angles $\vc{q}_t^{target}$ of the proportional-derivative (PD) controllers, deviating from the joint angles in the reference motion: 
\begin{equation}
    \label{eq:action}
    \vc{q}^{target}_{t} = \hat{\vc{q}}_{t}(\phi) + \vc{a}_{t},
\end{equation}
where $\hat{\vc{q}}_{t}(\phi)$ is the corresponding joint position in the reference motion at the given phase $\phi$. Our reward function is designed to imitate the reference motion:
\begin{multline}
    \label{eqn:reward}
    r_{h}(\vc{s}_{h},\vc{a}_{h}) = w_{q}(\vc{q}- \hat{\vc{q}}(\phi)) + w_{c}(\vc{c} - \hat{\vc{c}}(\phi)) \\ 
    + w_{e}(\vc{e} - \hat{\vc{e}}(\phi)) 
    - w_{\tau}||\boldsymbol{\tau}||^{2}
\end{multline}

where $\hat{\vc{q}}$, $\hat{\vc{c}}$, and $\hat{\vc{e}}$ are the desired joint positions, COM positions, and end-effector positions from the reference motion data, respectively. The reward function also penalizes the magnitude of torque $\boldsymbol{\tau}$. We use the same weight $w_{q} = 5.0$, $w_{c} = 2.0$, $w_{e} = 0.5$, and $w_{\tau} = 0.005$ for all experiments. We also use early termination of the rollouts, if the agent's pelvis drops below a certain height or if the base rotates about any axis beyond a threshold, we end the rollout and re-initialize the state.

We exert random forces to the agent during policy training. Each random force has a magnitude uniformly sampled from $[0,800]\ N$ and a direction uniformly sampled from [-$\pi/2$,$\pi/2$], applied for $50$ milliseconds on the agent's pelvis in parallel to the ground. The maximum force magnitude induces a velocity change of roughly $0.6$m/sec. This magnitude of change in velocity is comparable to experiments found in literature such as \cite{Wang2014},\cite{Agarwal} and \cite{hof2010balance}. We also randomize the time when the force is applied within a gait cycle. Training in such a stochastic environment is crucial for reproducing the human motion. We represent a human policy as a multi-layered perceptron (MLP) neural network with two hidden layers of $128$ neurons each. The formulated MDP is trained with Proximal Policy Optimization (PPO) \cite{schulman2017proximal}.

\subsection{Parameter Identification} 
As described in the previous section, the policy $\pi_h$ is a function of vector $\mu$. So, once the policy is trained we can search over the parameter $\mu$ to identify the values which best explains reference motion data. To do this, we perform an optimization step using CMA-ES algorithm \cite{Hansen16a}. We use the same reward function as described in equation 2. In this optimization process, at the beginning of each trajectory we set the value of $\mu$, then generate the motion by executing the policy $\pi_h$, at the end of each generated trajectory the reward is computed using the equation 3. The goal is to maximize the sum of reward accumulated along the trajectory generated by the policy $\pi_h$.

\begin{equation}
\label{eqn:parid}
\mu^{*} = \arg\max_{\mu}^{(s,a) ~\pi} \sum_{t=0}^{T}(r_{h}^{t}(s_{h}^{t},a_{h}^{t}))
\end{equation}

\subsection{Error Threshold}

We iterate between policy optimization and parameter identification until the error, defined in equation \ref{eqn:errorthrehold}, is less than a predefined threshold $\kappa$.

\begin{equation}
\label{eqn:errorthrehold}
\epsilon = \sum_{t=0}^{T} ||q - \hat{q}|| + ||\tau - \hat{tau}|| + ||GRF - \hat{GRF}||)
\end{equation}

Here, $T$ is the time taken to complete one gait cycle, $q$ is the joint angles generated by the policy and $\hat{q}$ is the ground truth joint angles for that particular human subject. Similarly, $\tau$ is the joint moments and $GRF$ is the ground reaction forces (vector of dimension $R^3$). $\hat{\tau}$ and $\hat{GRF}$ are the ground truth joint moments and ground reaction forces measured in the real-world.

The procedure is outlined in algorithm \ref{alg:vahrehvah}.

\begin{algorithm}[!ht]
\caption{Algorithm}\label{alg:vahrehvah}
\begin{algorithmic}[1]
\FOR{human subject i:1...n}
\STATE \textbf{Input:} Dataset $D_{i}$ - Motion capture trajectory of Human-subject $i$, $\kappa$ - error threshold
\STATE Initialize CMA optimizer with generation size 8  
\STATE Initialize $\pi_{i}(a|s,\mu)$, $V_{i}(s)$ 
\WHILE{ $\epsilon > \kappa$} \label{line:trainingSetBegin}
\STATE Optimize policy $\pi$ using reward function \ref{eqn:reward}
\STATE Optimize $\mu^{*}$ using equation \ref{eqn:parid}
\STATE Compute error $\epsilon$ using \ref{eqn:errorthrehold}
\ENDWHILE \label{line:trainingSetEnd}
\ENDFOR
\RETURN{ $\mu^{*}$ and $\pi$ }
\end{algorithmic}
\end{algorithm}

\section{Results}
We trained walking policies for 5 human subjects (first 5 in table \ref{tab:data}). We do the following evaluation to validate our approach,
\begin{enumerate}
    \item Ablation study : How well does the joint angles, joint moments and ground reaction forces generated by our method compare to a baseline method which only uses RL? In other words, how much of a change does the second optimization step benefit the accuracy?
    We use Root-mean squared error as the metric for comparison.
    \item We highlight preliminary analysis on the generalizability of the approach by comparing the joint angles of 5 different human subjects.
\end{enumerate}

\subsection{Ablation Study}

\begin{figure}[!htb]
\centering
\setlength{\tabcolsep}{1pt}
\renewcommand{\arraystretch}{0.7}
 
  \includegraphics[width=0.4\textwidth,height=5cm]{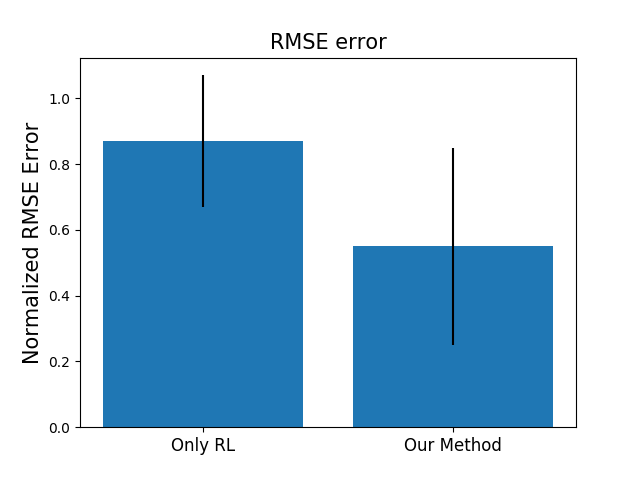}
   
\caption{RMSE error of our method compared to baseline approach which uses just RL to train the walking policy. We use 6 different initialization seeds for the policy, the bar plot indicates the RMSE error mean and standard deviation for the 6 policies for the same subject.}
\label{fig:Error}
\end{figure}

\begin{figure}
\centering
\includegraphics[width=0.8\linewidth]{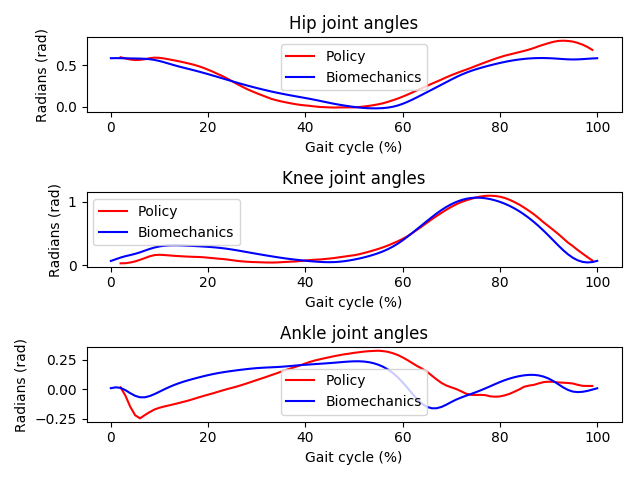}
\includegraphics[width=0.8\linewidth]{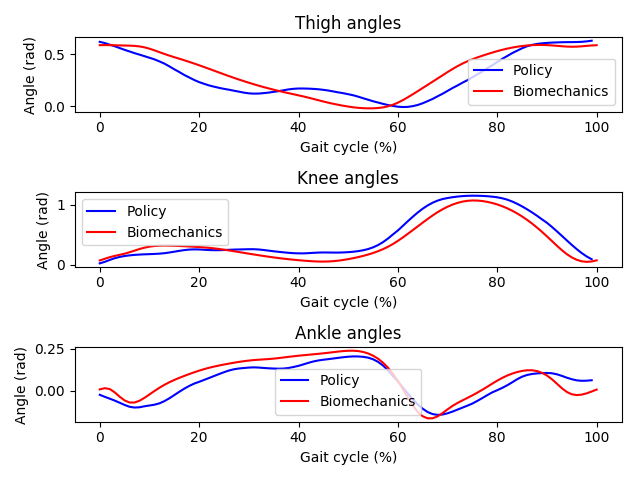}
\caption{Comparison of lower-limb joint angles during one gait cycle.  \textbf{Top:} Joint angles after RL step, the ankle joints are noticeably different although the curve profile is similar. \textbf{Bottom:} After CMA-ES optimization step, the joint angles match better with real-world data. 
} 
\label{fig:JAComparison}
\end{figure}

This ablation study is done only for one subject. In figure \ref{fig:Error}, we illustrate the benefit of our method, the additional optimization step has a clear effect on the RMSE error. Further, in figure \ref{fig:JAComparison}, we also show the lower-limb joint angles before and after the CMA-ES optimization step. The ankle joint in particular matches the real-world data more closely. This highlights the ability of control policies trained using RL to be robust to variation in the dynamical parameters.

\subsection{Joint moments, ground reaction forces and work loops}

\begin{figure}[!htb]
\centering
\setlength{\tabcolsep}{1pt}
\renewcommand{\arraystretch}{0.7}
 
  \includegraphics[width=0.4\textwidth,height=5cm]{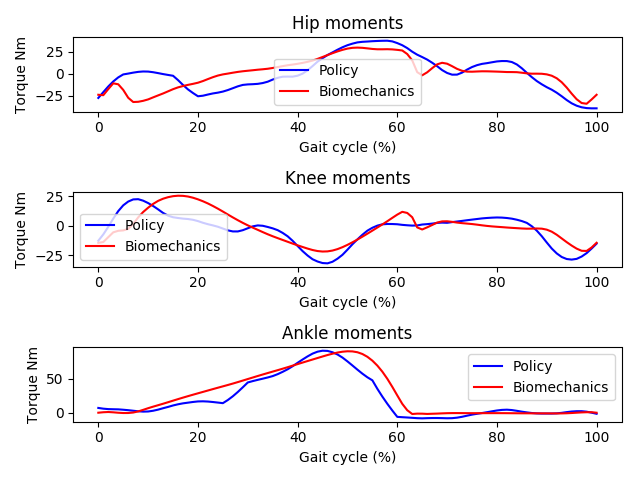}

\caption{Joint moments vs ground truth data.}
\label{fig:JM}
\end{figure}

\begin{figure}[!htpb]
\centering
\includegraphics[width=0.8\linewidth]{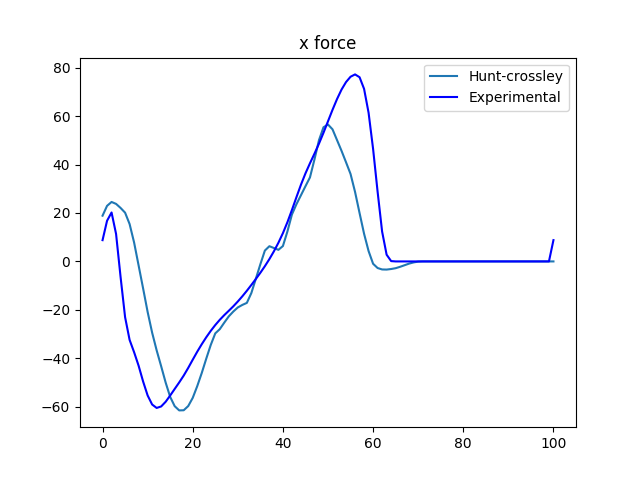}
\includegraphics[width=0.8\linewidth]{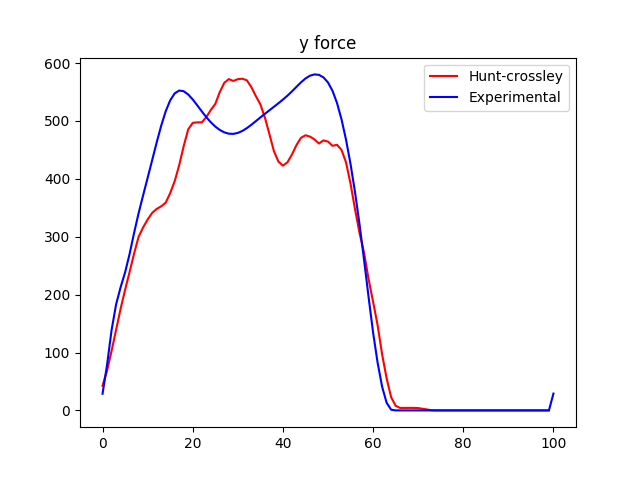}
\caption{Ground reaction forces \textbf{Top:} Tangential ground reaction force, its the force that propels a person forward while walking \textbf{Bottom:} Vertical ground reaction force, this is the normal force a person experiences while walking.
} 
\label{fig:grf}
\end{figure}

We also compare the joint moments and ground reaction forces generated by the policy to ground truth data by our method. The comparisons are illustrated in figures \ref{fig:JM} and \ref{fig:grf} respectively. 

In addition to this, we also compare the torque loop at the hip joint and compare it to the bio-mechanical data reported in \cite{Winter} (a long-standing gold standard). Torque loop is a plot of the torques generated at the hip joint in the y-axis and the joint angle in the x-axis during one-gait cycle. The arrows indicate the direction of movement and the points $[0,50,100]$\% of the gait cycle match fairly well. 

\begin{figure}[!htb]
\centering
\setlength{\tabcolsep}{1pt}
\renewcommand{\arraystretch}{0.7}
 
  \includegraphics[width=0.4\textwidth,height=5cm]{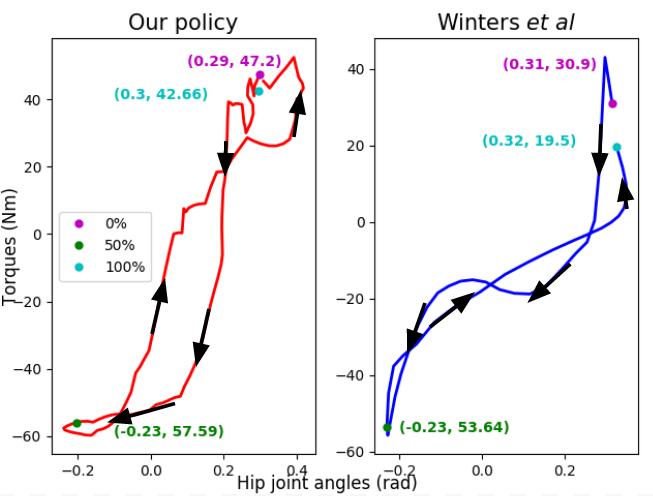}

\caption{Comparison of torque loops of a typical trajectory generated by our policy and human data reported by \cite{Winter} at the hip of stance leg during a gait cycle. The green dots indicate the start and the black dots indicate 50\% of the gait cycle. The arrows show the progression of the gait from 0\% to 100\%. }
\label{fig:workloop}
\end{figure}

\subsection{Results for five subjects}

\begin{figure}
\centering
\includegraphics[width=0.8\linewidth]{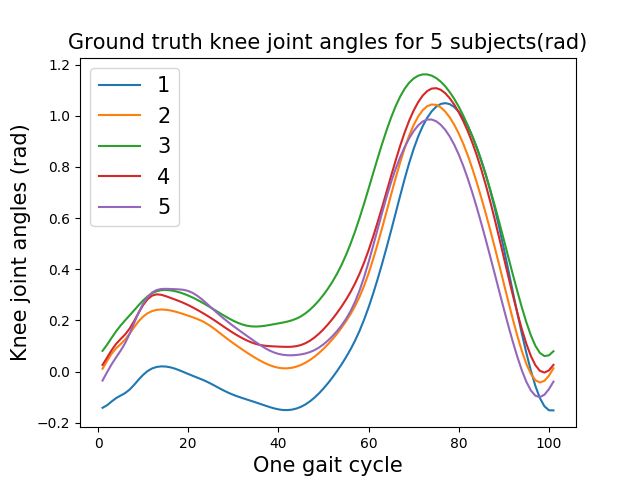}
\includegraphics[width=0.8\linewidth]{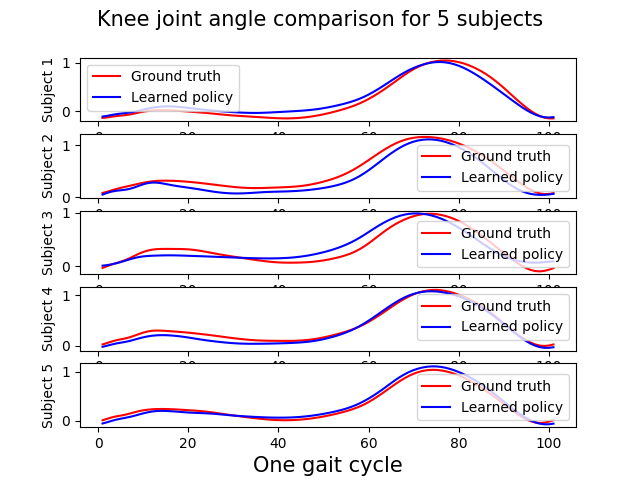}
\caption{Comparison of recovery performance when perturbation is applied at four different phases. \textbf{Top:} Comparison of stability region. \textbf{Bottom:} Comparison of COM velocity across five gait cycles. Perturbation is applied during the gait cycle 'p'. The increasing velocity after perturbation indicates that our policy is least effective at recovering when the perturbation occurs later in the swing phase.
} 
\label{fig:gen}
\end{figure}

Our approach generalizes well to different human subjects. We apply our approach to 5 different human subjects walking at 5 different speeds. Figure \ref{fig:gen} \textbf{(top)} illustrates the ground truth knee joint angle profiles for these 5 different human subjects. This shows the varied nature of the joint trajectories depending on speed and individual characteristics such as leg-length and mass. Our approach can capture these differences, as shown in figure \ref{fig:gen} \textbf{(bottom)}, the joint angles generated for the 5 different policies match well with the ground truth data.

\section{Discussion and Conclusion}
In this work, we have shown that our proposed method can generate 3D human walking motion that matches well with real-world data. Our ablation study indicates there is an improvement over existing RL methods to learn walking motion. There are plenty of questions we can try to answer in future steps:

\begin{itemize}
    \item First, we would like to test our algorithm on all 42 subjects in the open source dataset \cite{Fukuchi2018}.
    \item How well can we model a walking gait generated by a person with disability?
    \item How well can we model a walking gait generated by a person wearing an assistive device such as hip/ankle exoskeletons?
    \item Can we leverage this human model to learn interesting assistive strategies for an exoskeleton such as assistive walking to reduce metabolic cost of walking or to assist recovery when there is an external perturbation such as trips  or slips?
\end{itemize}





\bibliographystyle{IEEEtran}
\bibliography{sample}

\end{document}

%% file: defs.tex

\newcommand{\cmt}[1]{}
\newcommand{\visak}[1]{\textcolor{blue}{{Visak: #1}}}
\newcommand{\sehoon}[1]{\textcolor{magenta}{{Sehoon: #1}}}
\newcommand{\karen}[1]{\textcolor{red}{{Karen: #1}}}
\newcommand{\newtext}[1]{#1}
\newcommand{\original}[1]{\textcolor{magenta}{Original: #1}}
\newcommand{\eqnref}[1]{Equation~(\ref{eq:#1})}
\newcommand{\figref}[1]{Figure~\ref{fig:#1}}
\newcommand{\algref}[1]{Algorithm~\ref{alg:#1}}
\newcommand{\tabref}[1]{Table~\ref{tab:#1}}
\newcommand{\secref}[1]{Section~\ref{sec:#1}}

\long\def\ignorethis#1{}

\newcommand{\etal}{{\em{et~al.}\ }}
\newcommand{\eg}{e.g.\ }
\newcommand{\ie}{i.e.\ }

\newcommand{\figtodo}[1]{\framebox[0.8\columnwidth]{\rule{0pt}{1in}#1}}



\newcommand{\pdd}[3]{\ensuremath{\frac{\partial^2{#1}}{\partial{#2}\,\partial{#3}}}}

\newcommand{\mat}[1]{\ensuremath{\mathbf{#1}}}
\newcommand{\set}[1]{\ensuremath{\mathcal{#1}}}

\newcommand{\vc}[1]{\ensuremath{\mathbf{#1}}}
\newcommand{\vEndEff}{\ensuremath{\vc{d}}}
\newcommand{\vRelMove}{\ensuremath{\vc{r}}}
\newcommand{\sSet}{\ensuremath{S}}

\newcommand{\vControl}{\ensuremath{\vc{u}}}
\newcommand{\vPoint}{\ensuremath{\vc{p}}}
\newcommand{\sSpringCoef}{{\ensuremath{k_{s}}}}
\newcommand{\sDamperCoef}{{\ensuremath{k_{d}}}}
\newcommand{\vHandle}{\ensuremath{\vc{h}}}
\newcommand{\vForce}{\ensuremath{\vc{f}}}

\newcommand{\mTransChain}{\ensuremath{\vc{W}}}
\newcommand{\mRotateTrans}{\ensuremath{\vc{R}}}
\newcommand{\sJoint}{\ensuremath{q}}
\newcommand{\vJoint}{\ensuremath{\vc{q}}}
\newcommand{\mJoint}{\ensuremath{\vc{Q}}}
\newcommand{\mMass}{\ensuremath{\vc{M}}}
\newcommand{\sMass}{\ensuremath{{m}}}
\newcommand{\vGravity}{\ensuremath{\vc{g}}}
\newcommand{\vConstr}{\ensuremath{\vc{C}}}
\newcommand{\sConstr}{\ensuremath{C}}
\newcommand{\vCOM}{\ensuremath{\vc{x}}}
\newcommand{\sGeneralForce}[1]{\ensuremath{Q_{#1}}}
\newcommand{\vStateVar}{\ensuremath{\vc{y}}}
\newcommand{\vControlVar}{\ensuremath{\vc{u}}}
\newcommand{\tr}[1]{\ensuremath{\mathrm{tr}\left(#1\right)}}

%
%

\renewcommand{\choose}[2]{\ensuremath{\left(\begin{array}{c} #1 \\ #2 \end{array} \right )}}

\newcommand{\gauss}[3]{\ensuremath{\mathcal{N}(#1 | #2 ; #3)}}

\newcommand{\pctab}{\hspace{0.2in}}
\newenvironment{pseudocode} {\begin{center} \begin{minipage}{\textwidth}
                             \normalsize \vspace{-2\baselineskip} \begin{tabbing}
                             \pctab \= \pctab \= \pctab \= \pctab \=
                             \pctab \= \pctab \= \pctab \= \pctab \= \\}
                            {\end{tabbing} \vspace{-2\baselineskip}
                             \end{minipage} \end{center}}
\newenvironment{items}      {\begin{list}{$\bullet$}
                              {\setlength{\partopsep}{\parskip}
                                \setlength{\parsep}{\parskip}
                                \setlength{\topsep}{0pt}
                                \setlength{\itemsep}{0pt}
                                \settowidth{\labelwidth}{$\bullet$}
                                \setlength{\labelsep}{1ex}
                                \setlength{\leftmargin}{\labelwidth}
                                \addtolength{\leftmargin}{\labelsep}
                                }
                              }
                            {\end{list}}
\newcommand{\newfun}[3]{\noindent\vspace{0pt}\fbox{\begin{minipage}{3.3truein}\vspace{#1}~ {#3}~\vspace{12pt}\end{minipage}}\vspace{#2}}

\newcommand{\key}{\textbf}
\newcommand{\fun}{\textsc}

